\documentclass[runningheads]{llncs}
\usepackage[T1]{fontenc}
\usepackage{cleveref}
\usepackage{graphicx}
\usepackage{diagbox}
\usepackage{booktabs}
\usepackage{multirow}

\begin{document}
\title{Deep Learning-Based Fetal Lung Segmentation from Diffusion-weighted MRI Images and Lung Maturity Evaluation for Fetal Growth Restriction}
\author{}
\institute{}
%
\author{Zhennan Xiao\inst{1} \and Katharine Brudkiewicz\inst{1,2} \and Zhen Yuan \inst{1}\and Rosalind Aughwane\inst{2,3} \and Magdalena Sokolska\inst{3,4} \and Joanna Chappell\inst{1,3} \and Trevor Gaunt\inst{5} \and Anna L. David\inst{2,3} \and Andrew P. King\inst{1} \and Andrew Melbourne\inst{1}}
\authorrunning{Zhennan Xiao et al.}
\institute{School of Biomedical Engineering and Imaging Sciences, King's College London, London, UK \and Elizabeth Garrett Anderson Institute for Women’s Health, University College, London, UK 
\email{}\\
\url{}  
\and University College London Hospital NHS Foundation Trust, London, UK \and Department of Medical Physics and Biomedical Engineering, University College London Hospitals, London, UK \and Department of Radiology, University College London Hospitals NHS Foundation Trust, London, United Kingdom\\
}
\maketitle              
\begin{abstract}
Fetal lung maturity is a critical indicator for predicting neonatal outcomes and the need for post-natal intervention, especially for pregnancies affected by fetal growth restriction. Intra-voxel incoherent motion analysis has shown promising results for non-invasive assessment of fetal lung development, but its reliance on manual segmentation is time-consuming, thus limiting its clinical applicability. In this work, we present an automated lung maturity evaluation pipeline for diffusion-weighted magnetic resonance images that consists of a deep learning-based fetal lung segmentation model and a model-fitting lung maturity assessment. A 3D nnU-Net model was trained on manually segmented images selected from the baseline frames of 4D diffusion-weighted MRI scans. The segmentation model demonstrated robust performance, yielding a mean Dice coefficient of 82.14$\%$. Next, voxel-wise model fitting was performed based on both the nnU-Net-predicted and manual lung segmentations to quantify IVIM parameters reflecting tissue microstructure and perfusion. The results suggested no differences between the two. Our work shows that a fully automated pipeline is possible for supporting fetal lung maturity assessment and clinical decision-making.

\end{abstract}
\section{Introduction}
\label{Introduction}
Fetal growth restriction (FGR) refers to the clinical condition where a fetus fails to achieve its genetic growth potential \cite{aughwane2020placental}. It affects around 7-10\% of pregnancies worldwide \cite{lee2013national}. Typically, FGR is challenging to detect as it is often overlaps with the small for gestational age fetus (SGA fetuses). Without timely diagnosis, FGR may result in late stillbirth \cite{flouri2022placental}. In the absence of effective treatment, early delivery combined with postnatal care remains a common clinical strategy to improve fetal survival \cite{nawathe2018prophylaxis}. However, since the fetal lung is among the last organs to reach its functional maturity \cite{luo2008revisiting}, early delivery could potentially result in respiratory complications \cite{mills2014determination} due to lung immaturity. Therefore, accurate \textit{in utero} assessment of fetal lung maturity is vital to enable accurate perinatal decision-making and potential post-natal treatment strategies, thus improving the fetal survival rate.

In the current standard clinical workflow, biochemical and biophysical tests such as amniocentesis are the most commonly used methods to determine fetal lung maturity \cite{mills2014determination}. Amniocentesis measures the lecithin-sphingomyelin (L/S) ratio in amniotic fluid, with higher values indicating sufficient fetal lung development. However, those traditional methods are invasive and pose risks such as infection and preterm labour \cite{mills2014determination}. Imaging-based techniques, including ultrasound and magnetic resonance imaging (MRI), offer safer alternatives for fetal assessment \cite{kasprian2006mri}. While ultrasound is cheap and widely accessible, its diagnostic accuracy is dependent upon image quality and operator expertise \cite{flouri2022placental}. In contrast, MRI provides better soft tissue contrast and multiplanar views \cite{florkow2022magnetic}, making it increasingly valuable for evaluating fetal lung development \cite{kertes2025ivim}. Furthermore, Intra-voxel Incoherent Motion (IVIM) is a non-invasive and contrast agent-free \cite{paschoal2022contrast} MRI technique that separates molecular water diffusion from microvascular perfusion effects on MRI signal intensity \cite{le2019can}, useful for lung maturity assessment in fetal imaging \cite{kertes2025ivim}. However, manual segmentation of lung regions for later model-fitting is time-consuming. 

In this study, we employed the nnU-Net framework \cite{isensee2021nnu} for automated \textit{in utero} fetal lung segmentation from diffusion-weighted (DWI) MRI images. By comparing the IVIM-derived quantitative parameters (e.g. tissue perfusion and diffusion coefficients) from manual and automated segmentations, we demonstrate that deep learning can reliably replace manual delineation, enhancing the efficiency of IVIM-based analysis. In summary, our study establishes a practical and reliable solution for fetal lung segmentation and quantitative analysis using DWI MRI images.

\section{Methods}
\label{Methods}

\subsection{Automatic Lung Segmentation based on nnU-Net}
\label{Automatic Lung Segmentation based on nnU-Net}
\begin{figure}[htp]
    \centering
    \includegraphics[width=0.75\textwidth]{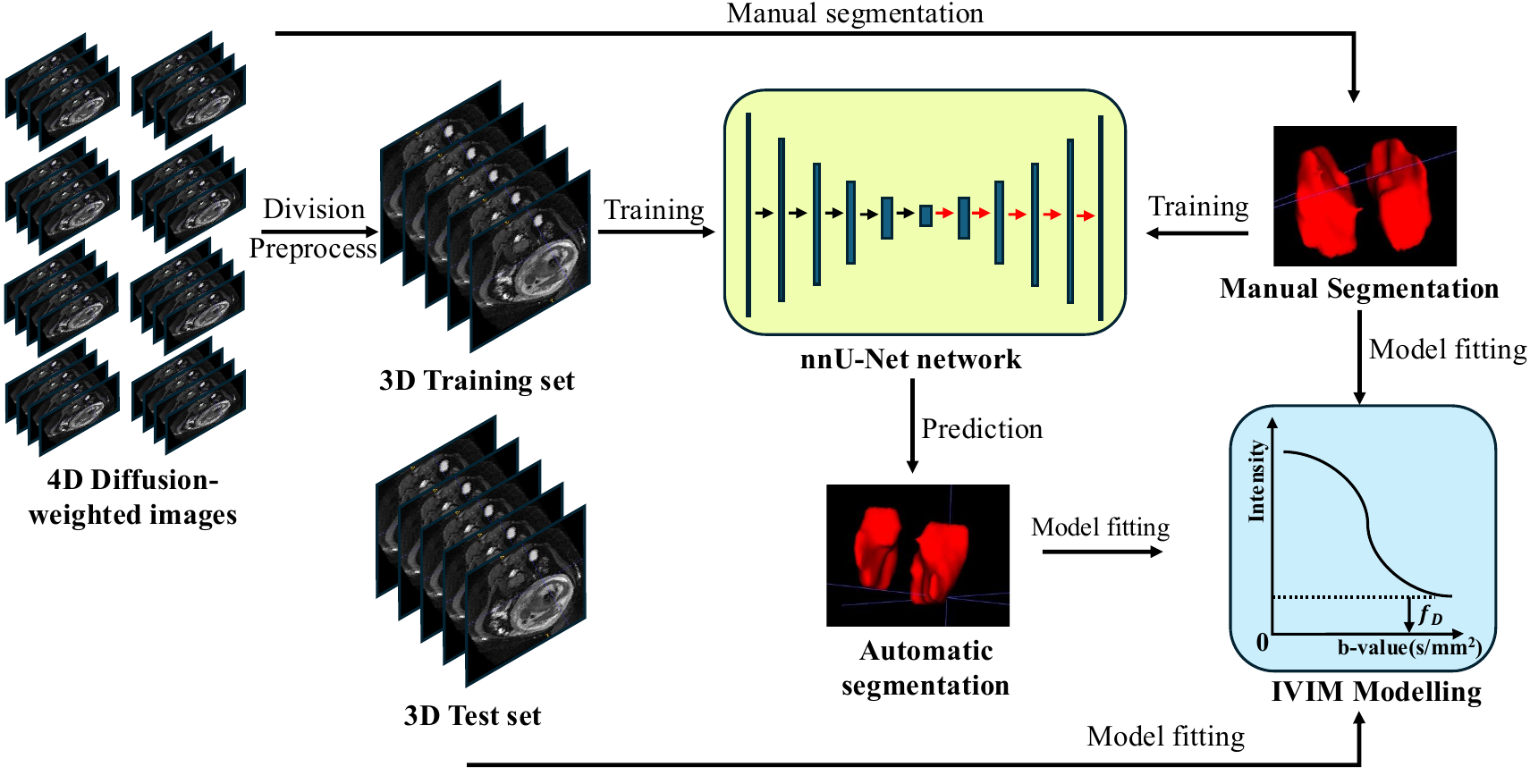}
    \caption{Overview of the proposed framework: manual segmentation of 4D DWI MRI images, data preprocessing, nnU-Net training for automated lung segmentation, and IVIM model fitting using both manual and automated masks to evaluate deep learning as a substitute for manual annotation.}
    \label{fig:img1}
\end{figure}

We adopted the nnU-Net framework \cite{isensee2021nnu} for automatic fetal lung segmentation from 3D DWI MRI volumes, as it has demonstrated its effectiveness in a range of datasets \cite{isensee2021nnu}\cite{yuan2025effect} and enables automated configuration optimisation. The default preprocessing procedures included resampling, normalisation, and patch-based data cropping. The proposed overall pipeline is presented in Fig. \ref{fig:img1}. 

\subsection{IVIM Modelling and Parameter Estimation}
The IVIM mathematical model describes the signal attenuation as:
\begin{equation}
    S(b)=S_0[fe^{-D^*b}+(1-f)e^{-Db}]
\end{equation}
where $b$ is the diffusion weighting (b-value), $S(b)$ is the signal intensity at that b-value, and $S_0$ is the signal intensity at $b=0$. The perfusion fraction $f$ represents the proportion of signal contributed by microvascular perfusion. $D^*$ denotes the pseudo-diffusion coefficient related to incoherent microvascular flow, while $D$ is the tissue diffusion coefficient reflecting standard Gaussian water diffusion \cite{le2019can}\cite{Flouri2020ivim}.

For each 4D dataset, we generated a single representative lung-volume segmentation using three different fusion methods to assess the impact of segmentation consensus on the final IVIM parameter estimation. These methods were: taking the intersection of all segmentations (OLP) to define the most conservative, consistently imaged 'core' lung volume; generating a binary mask of voxels presented in more than 50\% of the individual segmentations (AVG) as a robust majority-vote estimate; and taking the union of all segmentations (LC) to create the most inclusive contour capturing the full extent of lung motion \cite{yuan2025effect}. 
 
IVIM parameters were estimated using a two-step, voxel-wise least-squares fitting procedure in MATLAB, employing the Levenberg-Marquardt algorithm. First, the apparent diffusion coefficient (ADC) was measured by fitting a mono-exponential model to data from high b-values (\(b>100\;\mathrm{s/mm}^2\)) using: 
 \begin{equation}
    S(b)=S_0e^{-b\textrm{ADC}}
\end{equation}
The ADC parameter quantifies the rate of MR signal decay with increasing diffusion weighting and therefore serves as an indirect measure of tissue microstructural density. Since the ADC is only weakly dependent on low b-values, it can be prefitted independently and then be used to fix the parameter D in the biexponential IVIM model, improving the stability when solving for $f$ and $D^*$.

The spatial heterogeneity of the resulting parameter maps was assessed using the coefficient of variation (CV) and Shannon entropy. Entropy, a numerical measure of the randomness or unpredictability in the parameter distribution, was calculated as:
\begin{equation}
    H = -\sum_{i} p_i \log_2(p_i)
\end{equation}
where $p_i$ is the probability of the $i$-th parameter bin, obtained from a normalised histogram of the voxel-wise values within the segmentation mask. 

\subsection{FGR Classification using IVIM Parameters}
To identify a robust biomarker for FGR classification, a univariate analysis was performed on the various IVIM parameters derived from the manual segmentations of our training dataset, comparing the FGR and control cases. Among the tested parameters, the total lung volume (TLV) was the only one to show a statistically significant difference between the two groups ($p = 0.003$). Consequently, the observed-to-expected total lung volume ratio (oeTLV) was selected as the predictive feature for our classification model. 

The expected TLV was estimated using the $GA$ based model proposed by Cannie et al. \cite{Cannie2008}:
\begin{equation}
    Expected TLV = -0.0132 \cdot GA^3 + 1.14 \cdot GA^2 - 27.38 \cdot GA + 207.50
\end{equation}
where $GA$ is the gestational age in weeks. The observed-to-expected ratio was then calculated by dividing the measured TLV by the expected TLV.

The oeTLVs were standardised to Z-scores using the mean and standard deviation from the control cases within the manual training set. A Receiver-Operating-Characteristic analysis of this training set data then yielded an optimal Youden-index threshold. This threshold was subsequently applied to the independent test set to evaluate predictive accuracy on the automated segmentations. 

\section{Data Description and Experiments}
\label{Data Description and Experiments}
\subsection{Data Description and Preprocessing}
\begin{figure}[htp]
    \centering
    \includegraphics[width=0.75\textwidth]{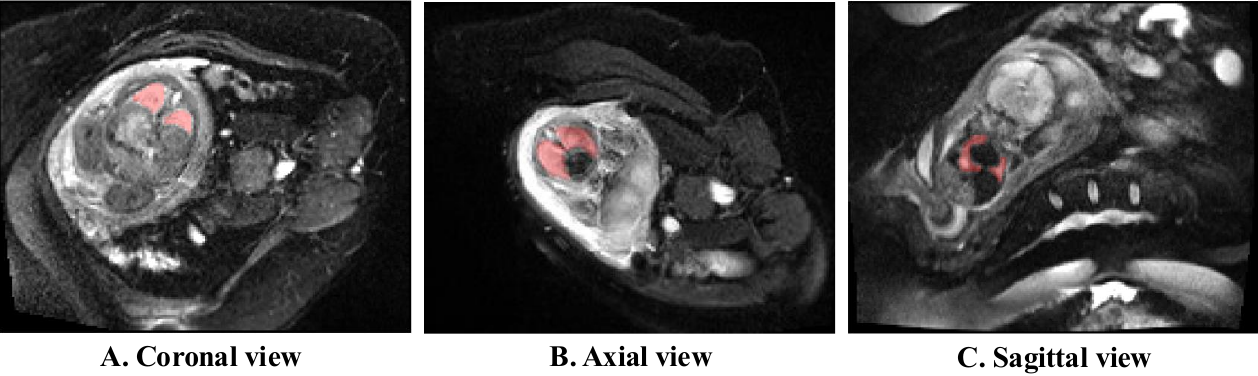}
    \caption{Representative fetal lung visibility in different views. (A) Coronal and (B) axial views show clear lung boundaries, while (C) sagittal view has poor definition, limiting its segmentation utility.}
    \label{fig:img2}
\end{figure}

We collected a total of 95 fetal 4D DWI MRI scans from 30 pregnant women (16 FGR pregnancies, 14 controls/normal pregnancies) at University College Hospital, London. The imaging orientation is defined relative to maternal anatomy (see Fig. \ref{fig:img2}) and there are 20 axial cases and 10 coronal cases in the dataset. The gestational age (GA) ranged from 20+0 to 36+0 weeks, with a median GA of 28+2 weeks overall (FGR: 28+2; control: 28+0). All scans were acquired using a 1.5T MRI system (Siemens Healthineers, Erlangen, Germany) with a spin-echo echo-planar imaging (SE-EPI) sequence. The protocol included multiple b-values (0–600 s/mm²) and diffusion directions to capture IVIM dynamics. Each 4D scan typically contained 7–10 temporal frames, each representing a 3D volume.

The primary imaging parameters were as follows: repetition time (TR) = 3700 ms, echo time (TE) = 73 ms, flip angle = 90°, slice thickness = 6 mm, in-plane resolution = 2.06 mm $\times$ 2.06 mm and inter-slice spacing = 7.2 mm. The acquisition matrix was 148 $\times$ 128. Manual segmentations were performed frame by frame using ITK-SNAP \cite{yushkevich2016itk} by two trained experts.

For model training, only the first timeframe (b = 0 s/mm²) was used, as it provided the clearest anatomical delineation with minimal diffusion attenuation. All images underwent the following preprocessing: (1) intensity values were rescaled to 0–255 and background was masked to zero, reducing inter-sample variance and artefacts \cite{kociolek2020does}; (2) N4 bias field correction was applied to mitigate scanner-related intensity inhomogeneity, which is particularly effective for fetal MRI where such artefacts may obscure anatomical boundaries \cite{tustison2010n4itk}; and (3) intensity histograms were computed to characterise global signal distributions. We also conducted strict data stratification to ensure the ratio between FGR and control was similar in training and test sets, and that there was no overlap of subjects or images between training and test sets.

\subsection{Segmentation Model Training}
The final dataset included 95 3D images. To reduce sampling bias and ensure balanced training and test sets, data were stratified by signal intensity statistics, GA, image orientation and group proportion as shown in Table \ref{tab1}.

\begin{table}[htbp]
\centering
\caption{Summary of dataset characteristics. GA: gestational age in weeks; A/C: ratio of axial to coronal images; F/C: ratio of FGR to control cases; I-mean and I-SD: mean and standard deviation of image intensity.}
\small
\begin{tabular}{>{\centering\arraybackslash}p{1.5cm}>{\centering\arraybackslash}p{1.5cm}>{\centering\arraybackslash}p{2cm}>{\centering\arraybackslash}p{1.5cm}>{\centering\arraybackslash}p{1.5cm}>{\centering\arraybackslash}p{1.5cm}>{\centering\arraybackslash}p{1.5cm}}
\toprule
\textbf{Set} & \textbf{No. of Images} & \textbf{Mean GA (weeks)} & \textbf{A/C Ratio} & \textbf{F/C Ratio} & \textbf{I-mean} & \textbf{I-SD} \\
\midrule
Training & 77 & 27.84 & 2.2:1 & 1.3:1 & 19.94 & 20.17 \\
Test     & 18 & 28.05 & 2:1   & 1:1   & 20.61 & 20.16 \\
\bottomrule
\end{tabular}
\label{tab1}
\end{table}

No subject contributed to both training and test sets to ensure strict data separation. All data were resampled to a voxel spacing of 7.20 × 2.07 × 2.07 mm and cropped into patches of 32 × 192 × 160 voxels. The 3D full-resolution nnU-Net pipeline was trained under default 5-fold cross-validation for 1,000 epochs per fold, using a batch size of 4 and a combined Dice–cross–entropy loss. The best model from each fold was selected based on validation Dice. Training was performed using PyTorch on NVIDIA A100 (40GB) GPUs.

\section{Results}
\label{Result}
\subsection{Evaluation of Automated Segmentation}
The nnU-Net model was evaluated on an independent test set of 18 3D images, each derived from one of 18 4D DWI MRI images across 6 fetuses (3 scans per fetus). For each fetus, the mean Dice coefficient and Hausdorff distance were computed across the three scans. Results are summarised in Table \ref{tab2}:

\begin{table}[htbp]
\centering
\caption{Quantitative evaluation of fetal lung segmentation performance on test set. Dice: Dice similarity coefficient; HD: Hausdorff Distance.}
\small
\begin{tabular}{>{\centering\arraybackslash}p{1.2cm}>{\centering\arraybackslash}p{2cm}>{\centering\arraybackslash}p{2.1cm}>{\centering\arraybackslash}p{2.1cm}>{\centering\arraybackslash}p{1.5cm}>{\centering\arraybackslash}p{1.7cm}}
\toprule
\textbf{Case} & \textbf{GA(weeks)} & \textbf{FGR/Control} & \textbf{Orientation} & \textbf{Dice (\%)} & \textbf{HD (mm)} \\
\midrule
1 & 30.00 & Control & Coronal & 86.08 & 8.74 \\
2 & 30.86 & FGR     & Axial   & 85.77 & 10.92 \\
3 & 28.43 & FGR     & Coronal & 81.75 & 11.47 \\
4 & 26.57 & Control & Axial   & 81.34 & 16.84 \\
5 & 25.00 & Control & Axial   & 79.39 & 15.93 \\
6 & 27.43 & FGR     & Axial   & 78.54 & 9.45  \\
\midrule
\textbf{Average} & 28.05 & -- & -- & 82.14 & 12.11 \\
\bottomrule
\end{tabular}
\label{tab2}
\end{table}

The nnU-Net achieved a mean Dice coefficient of 82.14$\%$ and a mean Hausdorff distance of 12.11 mm across the test set. As some Dice score distributions deviated from normality (per Shapiro–Wilk tests), Mann–Whitney U tests were used for all group comparisons. No significant differences were found between FGR and control fetuses (82.02±7.28$\%$ vs. 82.27±5.59$\%$; $p$=0.596), or between coronal and axial views (83.91±4.84$\%$ vs. 81.26±7.01$\%$; $p$=0.750). These findings suggest that the segmentation model performs robustly across clinical conditions and imaging orientations.

\begin{figure}[ht]
    \centering
    \includegraphics[width=0.75\textwidth]{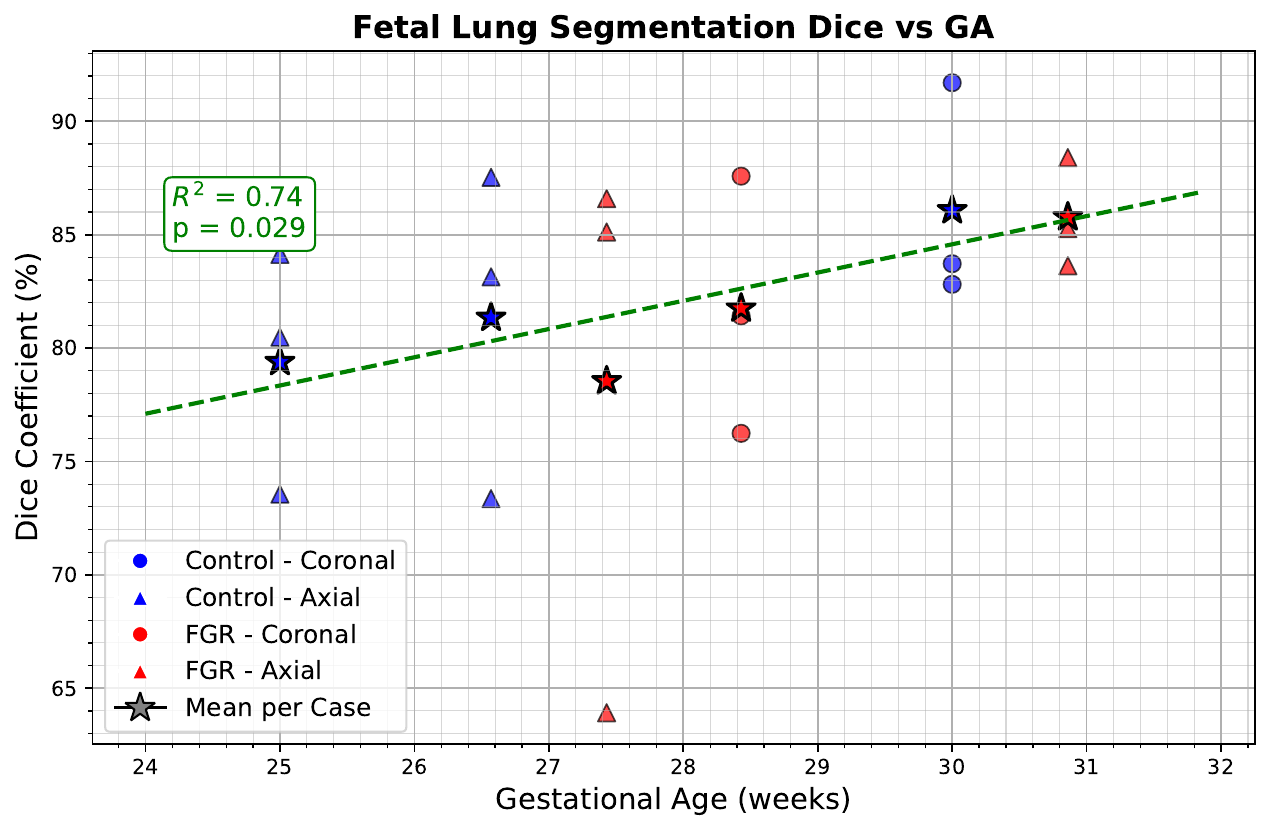}
    \caption{Relationship between GA and the Dice coefficient}
    \label{img:fig3}
\end{figure}

To explore factors affecting segmentation, we examined the relationship between GA and Dice scores (averaged per fetus). Linear regression revealed a significant positive correlation (R² = 0.74, $p$ = 0.029) as shown in Fig. \ref{img:fig3}, suggesting improved segmentation accuracy with advancing gestation, likely due to enhanced anatomical visibility and larger lung volume in later stages of pregnancy.

\vspace*{-0.5em}
\vspace*{-0.5em}

\subsection{Comparison of Manual and Automated IVIM Parameters}
Paired t-tests on each IVIM parameter (Volume, $S_0$, $f$, $D^*$, $ADC$, Residual) revealed no significant differences between manual and automatic segmentations for any fusion strategy (all $p\geq 0.1603$; Table \ref{tab3}). Similarly, there were no significant differences found for the variability within the parameter maps metrics (all $p\geq 0.0851$). Among the three fusion strategies, AVG yielded the highest average p-value across all metrics ($\bar{p}=0.5643$), followed by LC ($\bar{p}=0.4923$) and OLP ($\bar{p}=0.4315$). 

\vspace*{-1em}
\begin{table}[htbp]
\centering
\caption{Paired t-test p-values comparing Manual vs. Automatic segmentations for mean parameters and intra-mask variability metrics.}
\small
\setlength{\tabcolsep}{3pt}
\begin{tabular}{
    >{\centering\arraybackslash}p{2.2cm} 
    >{\centering\arraybackslash}p{0.9cm} 
    >{\centering\arraybackslash}p{0.9cm} 
    >{\centering\arraybackslash}p{0.9cm} 
    @{\hspace{8pt}} 
    >{\centering\arraybackslash}p{2.5cm} 
    >{\centering\arraybackslash}p{0.9cm} 
    >{\centering\arraybackslash}p{0.9cm} 
    >{\centering\arraybackslash}p{0.9cm}
}
\toprule
\multicolumn{4}{c}{\textbf{Mean Parameter p-values}} & \multicolumn{4}{c}{\textbf{Intra-Mask Variability p-values}} \\
\cmidrule(lr){1-4} \cmidrule(lr){5-8}
\textbf{Parameter} & \textbf{AVG} & \textbf{OLP} & \textbf{LC} & \textbf{Variability} & \textbf{AVG} & \textbf{OLP} & \textbf{LC} \\
\midrule
Volume & 0.6181 & 0.4394 & 0.5334 & $S_0^{cv}$ & 0.5907 & 0.3821 & 0.6355 \\
S0 & 0.9770 & 0.4987 & 0.5068 & $f^{cv}$ & 0.3933 & 0.6804 & 0.6518 \\
f & 0.8924 & 0.6556 & 0.7055 & $D^{*cv}$ & 0.4222 & 0.3133 & 0.2038 \\
D* & 0.1810 & 0.3748 & 0.1603 & $ADC_{cv}$ & 0.5315 & 0.3049 & 0.7741 \\
ADC & 0.1963 & 0.3936 & 0.4396 & $f_{\mathrm{entropy}}$ & 0.8121 & 0.3766 & 0.9619\\
Residual & 0.5631 & 0.4335 & 0.2276 & $D^*_{\mathrm{entropy}}$ & 0.7570 & 0.0851 & 0.2745 \\
& & & & $ADC_{\mathrm{entropy}}$ & 0.4017 & 0.6709 & 0.3256 \\
\bottomrule
\end{tabular}
\label{tab3}
\end{table}
\vspace*{-0.5em}

To assess the consistency of IVIM parameter estimation, we computed the inter-subject coefficient of variation (CV) for each IVIM parameter, separately for FGR and control cases (Table \ref{tab4}). The FGR cohort consistently demonstrated markedly higher subject-to-subject variability across most IVIM parameters compared to the control cohort, regardless of the segmentation or fusion method used. To quantify the agreement in measurement consistency between manual and automated methods, the mean absolute percentage difference in inter-subject CV was calculated. The AVG fusion method yielded the lowest overall mean difference of 28.4\%, followed by OLP (33.3\%) and LC (55.8\%). The high overall mean for LC fusion was influenced by the large differences in the FGR group for $f$ (+280.9\%) and ADC (+187.3\%) parameters. Notably, the OLP fusion had the lowest mean difference within the control group alone (14.8\%). The automated AVG method in some cases also improved measurement consistency, reducing the inter-subject CV for ADC in controls by 68.1\% compared to the manual method.

\vspace*{-1em}
\begin{table}[htbp]
\centering
\caption{Intra-group Coefficient of Variation (CV) for IVIM Parameters, comparing fusion strategies, Manual vs Automatic segmentations and control vs FGR.}
\scriptsize 
\begin{tabular}{
    >{\centering\arraybackslash}p{1.5cm} | 
    >{\centering\arraybackslash}p{0.8cm} >{\centering\arraybackslash}p{0.8cm} | 
    >{\centering\arraybackslash}p{0.8cm} >{\centering\arraybackslash}p{0.8cm} | 
    >{\centering\arraybackslash}p{0.8cm} >{\centering\arraybackslash}p{0.8cm} | 
    >{\centering\arraybackslash}p{0.8cm} >{\centering\arraybackslash}p{0.8cm} | 
    >{\centering\arraybackslash}p{0.8cm} >{\centering\arraybackslash}p{0.8cm} | 
    >{\centering\arraybackslash}p{0.8cm} >{\centering\arraybackslash}p{0.8cm}
}
\toprule
& \multicolumn{4}{c|}{\textbf{AVG Fusion}} 
& \multicolumn{4}{c|}{\textbf{OLP Fusion}} 
& \multicolumn{4}{c}{\textbf{LC Fusion}} \\
\cmidrule(lr){2-5} \cmidrule(lr){6-9} \cmidrule(lr){10-13}
& \multicolumn{2}{c|}{\textbf{Manual}} & \multicolumn{2}{c|}{\textbf{Automatic}} 
& \multicolumn{2}{c|}{\textbf{Manual}} & \multicolumn{2}{c|}{\textbf{Automatic}} 
& \multicolumn{2}{c|}{\textbf{Manual}} & \multicolumn{2}{c}{\textbf{Automatic}} \\
\cmidrule(lr){2-3} \cmidrule(lr){4-5} \cmidrule(lr){6-7} \cmidrule(lr){8-9} \cmidrule(lr){10-11} \cmidrule(lr){12-13}
\textbf{Parameter} & \textbf{Ctrl} & \textbf{FGR} & \textbf{Ctrl} & \textbf{FGR}
& \textbf{Ctrl} & \textbf{FGR} & \textbf{Ctrl} & \textbf{FGR}
& \textbf{Ctrl} & \textbf{FGR} & \textbf{Ctrl} & \textbf{FGR} \\
\midrule
Volume   & 0.3142 & 0.2399 & 0.3147 & 0.3280 & 0.5984 & 0.1124 & 0.7712 & 0.3200 & 0.1425 & 0.4109 & 0.1141 & 0.5157 \\
S$_0$      & 0.1374 & 0.3371 & 0.1579 & 0.3099 & 0.1386 & 0.3545 & 0.1681 & 0.3063 & 0.1708 & 0.3752 & 0.1849 & 0.3122 \\
$f$      & 0.0837 & 0.1337 & 0.1383 & 0.1138 & 0.1091 & 0.1749 & 0.1322 & 0.1000 & 0.0604 & 0.0628 & 0.0784 & 0.2434 \\
D$^*$    & 0.1137 & 0.1429 & 0.1706 & 0.2361 & 0.1761 & 0.1531 & 0.1664 & 0.1886 & 0.0394 & 0.1288 & 0.0361 & 0.1091 \\
ADC      & 0.0320 & 0.0558 & 0.0102 & 0.0482 & 0.0161 & 0.0873 & 0.0166 & 0.0491 & 0.0097 & 0.0300 & 0.0168 & 0.0862 \\
Residual & 0.4102 & 0.8862 & 0.3957 & 0.8871 & 0.4641 & 0.8998 & 0.4123 & 0.9188 & 0.3661 & 0.9519 & 0.3720 & 0.9207 \\
\bottomrule
\end{tabular}
\label{tab4}
\end{table}
\vspace*{-1em}

\subsection{Fetal Growth Restriction Classification}
On the 23‐case training set, oeTLV achieved an AUC of 0.9924, and the Youden‐index threshold was 3.751. When applied to the 6‐case test set, this threshold perfectly separated FGR and control cases (TN=3, FP=0; FN=0, TP=3) with a 100\% test accuracy.

\section{Discussion}
\label{Discussion}

In this work, we proposed a pipeline for automated assessment of fetal lung maturity from DWI MRI images. Results show that deep learning enables accurate fetal lung segmentation on DWI MRI images, supporting its potential for maturity assessment. No performance difference was observed across image orientations or clinical conditions (FGR). However, segmentation accuracy increased with GA, likely due to improved anatomical visibility and larger lung volume in later pregnancy.

We validated our automated segmentations for quantitative analysis and found no systemic bias. Paired t-tests revealed no significant differences between automated and manual masks for either the mean IVIM parameter values (all $p\geq 0.1603$) or their intra-mask variability (all $p\geq 0.0851$), confirming that the automated pipeline reliably captures both the magnitude and internal distribution of IVIM parameters. Among the fusion strategies, AVG fusion demonstrated the most consistently robust performance, combining a lack of significant systemic bias with the highest overall fidelity in inter-subject measurement reliability. While LC and OLP had high performance for preserving intra-mask characteristics or analysing control cohorts, their reliability was less consistent across the full spectrum of parameters. This makes AVG the preferred strategy in producing automated results that are broadly comparable to the manual delineation.

Our oeTLV-based model successfully classified FGR cases, confirming known lung volume reductions in FGR. However, volume alone does not reflect functional maturity. Since not all FGR fetuses suffer pulmonary compromise, there is a clear clinical need for functional biomarkers derived from microstructural and pathological information. Our pipeline provides a robust framework to develop such markers from IVIM modelling, including diffusion or perfusion parameters. Once validated against clinical outcomes like respiratory distress syndrome or bronchopulmonary dysplasia, these could enable personalised risk stratification and guide perinatal decision-making. Our work, therefore, establishes the feasibility of large-scale, automated analysis for this future research.

Despite these promising findings, several limitations should be noted. First, the model was trained on a small, single-centre dataset with limited protocol variability, and its generalisability to other scanners or populations remains to be validated. Furthermore, the relatively large Hausdorff distance (12 mm, Table \ref{tab2}) suggests occasional boundary mismatches, likely due to irregular lung shapes and low contrast in certain slices. Besides, the current approach neglects inter-frame fetal motion, potentially limiting its temporal coherence, which could be mitigated by slice-to-volume reconstruction techniques \cite{davidson2022motion}. Addressing these limitations through multi-centre validation, motion-robust processing, and spatiotemporal modelling will be essential to improve clinical readiness.

\vspace*{-0.5em}
\section{Conclusions}
\label{Conclusions}
We proposed a deep learning-based pipeline for fetal lung segmentation and maturity assessment from diffusion-weighted MRI. Using nnU-Net, we achieved accurate automated segmentations that enabled reliable voxel-wise IVIM modelling of the fetal lung. Our results support the feasibility of automated lung segmentation and suggest its clinical potential, particularly for evaluating fetuses complicated by FGR. Future work should focus on larger multi-centre datasets, robust validation, and improved motion compensation strategies.
\vspace*{-0.5em}


\bibliography{mybibfile.bib}  

\begin{thebibliography}{10}
\providecommand{\url}[1]{\texttt{#1}}
\providecommand{\urlprefix}{URL }
\providecommand{\doi}[1]{https://doi.org/#1}

\bibitem{aughwane2020placental}
Aughwane, R., Ingram, E., Johnstone, E.D., Salomon, L.J., David, A.L., Melbourne, A.: Placental mri and its application to fetal intervention. Prenatal diagnosis  \textbf{40}(1),  38--48 (2020)

\bibitem{Cannie2008}
Cannie, M.M., Jani, J.C., Van~Kerkhove, F., Meerschaert, J., De~Keyzer, F., Lewi, L., Deprest, J.A., Dymarkowski, S.: Fetal body volume at mr imaging to quantify total fetal lung volume: Normal ranges. Radiology  \textbf{247}(1),  197--203 (Apr 2008), \url{https://doi.org/10.1148/radiol.2471070682}

\bibitem{davidson2022motion}
Davidson, J., Uus, A., Egloff, A., Van~Poppel, M., Matthew, J., Steinweg, J., Deprez, M., Aertsen, M., Deprest, J., Rutherford, M.: Motion corrected fetal body magnetic resonance imaging provides reliable 3d lung volumes in normal and abnormal fetuses. Prenatal Diagnosis  \textbf{42}(5),  628--635 (2022)

\bibitem{florkow2022magnetic}
Florkow, M.C., Willemsen, K., Mascarenhas, V.V., Oei, E.H., van Stralen, M., Seevinck, P.R.: Magnetic resonance imaging versus computed tomography for three-dimensional bone imaging of musculoskeletal pathologies: a review. Journal of Magnetic Resonance Imaging  \textbf{56}(1),  11--34 (2022)

\bibitem{flouri2022placental}
Flouri, D., Darby, J.R., Holman, S.L., Cho, S.K., Dimasi, C.G., Perumal, S.R., Ourselin, S., Aughwane, R., Mufti, N., Macgowan, C.K., et~al.: Placental mri predicts fetal oxygenation and growth rates in sheep and human pregnancy. Advanced Science  \textbf{9}(30),  2203738 (2022)

\bibitem{Flouri2020ivim}
Flouri, D., Owen, D., Aughwane, R., Mufti, N., Maksym, K., Sokolska, M., Kendall, G., Bainbridge, A., Atkinson, D., Vercauteren, T., Ourselin, S., David, A.L., Melbourne, A.: Improved fetal blood oxygenation and placental estimated measurements of diffusion-weighted mri using data-driven bayesian modeling. Magnetic resonance in medicine  \textbf{83},  2160--2172 (Jun 2020)

\bibitem{isensee2021nnu}
Isensee, F., Jaeger, P.F., Kohl, S.A., Petersen, J., Maier-Hein, K.H.: nnu-net: a self-configuring method for deep learning-based biomedical image segmentation. Nature methods  \textbf{18}(2),  203--211 (2021)

\bibitem{kasprian2006mri}
Kasprian, G., Balassy, C., Brugger, P.C., Prayer, D.: Mri of normal and pathological fetal lung development. European Journal of Radiology  \textbf{57}(2),  261--270 (2006)

\bibitem{kertes2025ivim}
Kertes, N., Zaffrani-Reznikov, Y., Afacan, O., Kurugol, S., Warfield, S.K., Freiman, M.: Ivim-morph: Motion-compensated quantitative intra-voxel incoherent motion (ivim) analysis for functional fetal lung maturity assessment from diffusion-weighted mri data. Medical Image Analysis  \textbf{101},  103445 (2025)

\bibitem{kociolek2020does}
Kocio{\l}ek, M., Strzelecki, M., Obuchowicz, R.: Does image normalization and intensity resolution impact texture classification? Computerized Medical Imaging and Graphics  \textbf{81},  101716 (2020)

\bibitem{le2019can}
Le~Bihan, D.: What can we see with ivim mri? Neuroimage  \textbf{187},  56--67 (2019)

\bibitem{lee2013national}
Lee, A.C., Katz, J., Blencowe, H., Cousens, S., Kozuki, N., Vogel, J.P., Adair, L., Baqui, A.H., Bhutta, Z.A., Caulfield, L.E., et~al.: National and regional estimates of term and preterm babies born small for gestational age in 138 low-income and middle-income countries in 2010. The Lancet global health  \textbf{1}(1),  e26--e36 (2013)

\bibitem{luo2008revisiting}
Luo, G., Norwitz, E.R.: Revisiting amniocentesis for fetal lung maturity after 36 weeks’ gestation. Reviews in Obstetrics and Gynecology  \textbf{1}(2), ~61 (2008)

\bibitem{mills2014determination}
Mills, M., Winter, T.C., Kennedy, A.M., Woodward, P.J.: Determination of fetal lung maturity using magnetic resonance imaging signal intensity measurements. Ultrasound quarterly  \textbf{30}(1),  61--67 (2014)

\bibitem{nawathe2018prophylaxis}
Nawathe, A., David, A.L.: Prophylaxis and treatment of foetal growth restriction. Best Practice \& Research Clinical Obstetrics \& Gynaecology  \textbf{49},  66--78 (2018)

\bibitem{paschoal2022contrast}
Paschoal, A.M., Secchinatto, K.F., da~Silva, P.H.R., Zotin, M.C.Z., Dos~Santos, A.C., Viswanathan, A., Pontes-Neto, O.M., Leoni, R.F.: Contrast-agent-free state-of-the-art mri on cerebral small vessel disease—part 1. asl, ivim, and cvr. NMR in Biomedicine  \textbf{35}(8),  e4742 (2022)

\bibitem{tustison2010n4itk}
Tustison, N.J., Avants, B.B., Cook, P.A., Zheng, Y., Egan, A., Yushkevich, P.A., Gee, J.C.: N4itk: improved n3 bias correction. IEEE transactions on medical imaging  \textbf{29}(6),  1310--1320 (2010)

\bibitem{yuan2025effect}
Yuan, Z., Schabel, M.C., David, A.L., Roberts, V.H., Melbourne, A.: The effect of deep learning segmentation on placental t2* estimation. In: 2025 IEEE 22nd International Symposium on Biomedical Imaging (ISBI). pp.~1--5. IEEE (2025)

\bibitem{yushkevich2016itk}
Yushkevich, P.A., Gao, Y., Gerig, G.: Itk-snap: An interactive tool for semi-automatic segmentation of multi-modality biomedical images. In: 2016 38th annual international conference of the IEEE engineering in medicine and biology society (EMBC). pp. 3342--3345. IEEE (2016)

\end{thebibliography}

\end{document}